\RequirePackage{amsmath,mathtools,amsthm}

\documentclass[10pt]{iopart}

\newif\ifexternalized

\externalizedtrue

\newcommand{\De}{\;\coloneqq\;}
\newcommand{\Prox}[1]{\operatorname{prox}_{#1}}

\usepackage[square,sort&compress,numbers]{natbib}

\usepackage[english]{babel}
\usepackage[autostyle, english=british]{csquotes} 

\usepackage{iopams}  
\usepackage{amssymb}
\usepackage{color}
\usepackage{algorithm}
\usepackage[noend]{algpseudocode}

\usepackage{tikz}

\usetikzlibrary{external}
\tikzset{external/system call= {pdflatex \tikzexternalcheckshellescape -halt-on-error -interaction=batchmode -jobname "\image" "\texsource"}} 

\usetikzlibrary{spy}
\usepackage{pgfplots}
\pgfplotsset{compat=1.13}
\newcommand{\DrawVText}[3]{\draw[color=black] (#1,#2) node [anchor=center] {\rotatebox{90}{#3}};}%
\newcommand{\DrawText}[3]{\draw[color=black] (#1,#2) node [anchor=center] {{#3}};}%

\def\PicWidth{2.8cm}%
\def\KerWidth{1.2cm}%
\def\ZoomWidth{1.4cm}%
\def\MagnifyingFactor{3.5}%
\def\PicDistance{0.3cm}

\def\PosXa{0cm}
\def\PosXb{4.2cm}
\def\PosXc{8.4cm}
\def\PosYa{-0cm}
\def\PosYb{-3.5cm}
\def\PosYc{-7.0cm}
\def\PosYd{-10.5cm}

\newcommand{\PlotSpy}[6]{%
    \draw (#3, #4) node [anchor=south west] {\fbox{\includegraphics[width=\PicWidth]{#2}}};%
    \spy [black, draw, width=\ZoomWidth, height=\ZoomWidth, magnification=\MagnifyingFactor, connect spies] on (#3+#5,#4+#6) in node [anchor=south west] at (#3 + \PicWidth, #4-0.2cm);
    \draw (#3+1.5cm, #4-.1cm) node [anchor=center] {\footnotesize #1};}%
\newcommand{\PlotKernelSpy}[7]{%
    \PlotSpy{#1}{#2}{#4}{#5}{#6}{#7}
    \draw (#4+\PicWidth - .1cm, #5+\PicWidth + .4cm) node [anchor=north west] {\fbox{\includegraphics[height=\KerWidth]{#3}}};}%

\newcommand{\Argmin}[2]{\operatorname{arg} \underset{#1}{\min} \;#2}

\newcommand{\R}{\mathbb R}

\newcommand{\tv}{\mathrm{TV}}
\newcommand{\dtv}{\mathrm{dTV}}
\newcommand{\grad}{\nabla}
\newcommand{\Obj}{\Psi}
\newcommand{{\DataFid}}{\mathcal{D}}
\newcommand{\Ru}{\RegFun_u}
\newcommand{\Rk}{\RegFun_k}

\def\stepsize{\tau}

\def\spaceLarge{U}

\def\spaceKernel{K}

\def\Fourier{\mF}
\def\RegParam{\lambda}
\def\Simplex{\mathbb S}

\def\InertParamA{\alpha}
\def\iter{t}
\def\RegFun{\mathcal R}

\def\mA{\mathbf A}
\def\mB{\mathbf B}
\def\mC{\mathbf C}
\def\mF{\mathbf F}
\def\mH{\mathbf H}
\def\mI{\mathbf I}
\def\mJ{\mathbf J}
\def\mP{\mathbf P}
\def\mS{\mathbf S}

\theoremstyle{definition}
\newtheorem*{definition*}{Definition}

\def\DataGTdisk{\texttt{groundtruth\_disk}}
\def\DataGTgaussian{\texttt{groundtruth\_Gaussian}}
\def\DataTreesA{\texttt{environment1}}
\def\DataTreesB{\texttt{environment2}}
\def\DataUrbanA{\texttt{urban1}}
\def\DataUrbanB{\texttt{urban2}}

\usepackage[hyphens]{url}
\usepackage[breaklinks=true]{hyperref}
    
\newcommand{\rev}[1]{#1}

\begin{document}
\title[Blind Image Fusion for Hyperspectral Imaging]{Blind Image Fusion for Hyperspectral Imaging with the Directional Total Variation}

\author{Leon~Bungert$^1$, David~A.~Coomes$^2$, Matthias~J.~Ehrhardt$^3$, Jennifer~Rasch$^4$, Rafael~Reisenhofer$^5$ and Carola-Bibiane~Sch\"onlieb$^3$} 

\address{$^1$ Department of Mathematics, Friedrich-Alexander University Erlangen-N\"urnberg, Cauerstr. 11, 91058 Erlangen, Germany}
\address{$^2$ Forest Ecology and Conservation Group, Department of Plant Sciences, University of Cambridge, Cambridge CB2 3EA, United Kingdom}
\address{$^3$ Department for Applied Mathematics and Theoretical Physics, University of Cambridge, Cambridge CB3 0WA, United Kingdom}
\address{$^4$ Fraunhofer Heinrich Hertz Institute, Einsteinufer 37, 10587 Berlin, Germany}
\address{$^5$ University of Bremen, Fachbereich 3, Postfach 330440, 28334 Bremen, Germany}

\ead{m.j.ehrhardt@damtp.cam.ac.uk}

\begin{abstract}
Hyperspectral imaging is a cutting-edge type of remote sensing used for mapping vegetation properties, rock minerals and other materials. A major drawback of hyperspectral imaging devices is their intrinsic low spatial resolution. In this paper, we propose a method for increasing the spatial resolution of a hyperspectral image by fusing it with an image of higher spatial resolution that was obtained with a different imaging modality. This is accomplished by solving a variational problem in which the regularization functional is the directional total variation. To accommodate for possible mis-registrations between the two images, we consider a non-convex blind super-resolution problem where both a fused image and the corresponding convolution kernel are estimated. Using this approach, our model can realign the given images if needed. Our experimental results indicate that the non-convexity is negligible in practice and that reliable solutions can be computed using a variety of different optimization algorithms. Numerical results on real remote sensing data from plant sciences and urban monitoring show the potential of the proposed method and suggests that it is robust with respect to the regularization parameters, mis-registration and the shape of the kernel.
\end{abstract}

\vspace{2pc}

\submitto{\IP}

\section{Introduction}
Hyperspectral imaging is an earth observation technique that measures the energy of light reflected off the earth's surface within many narrow spectral bands, from which reflectance curves can be calculated. The shape of these reflectance curves provides information about the chemical and physical properties of materials such as minerals in rocks, vegetation, synthetic materials and water, which allows these materials to be classified and mapped remotely (see \cite{Ustin2004,Homolova2013}, for instance). 

Usually, hyperspectral images of the earth are being recorded by imaging devices that are mounted on planes or satellites. In either case, the speed of the carrier and narrowness of the spectral bands result in limited energy reaching the sensor for each spectral band. Therefore, most hyperspectral cameras have an intrinsic trade-off between spatial and spectral resolution \cite{Amro2011, Loncan2015}. A standard approach to overcoming this trade-off is to record a second image with low spectral but high spatial resolution (e.g. a photograph or a laser scan) \cite{Mowle1991} and to fuse these data together in order to create an image that has the high spectral resolution of the hyperspectral image and the high spatial resolution of the photograph \cite{Mowle1991, Ballester2006, Moeller2009, Padwick2010pansharpening, Amro2011, moller2012variational, Loncan2015, Simoes2015, Zhang2015pansharpening, Hou2016, Duran2017}. These techniques have become known as \emph{pansharpening} or \emph{image fusion}. The problem is illustrated by three example data sets in \fref{FIG:GOAL}.
{%
\newcommand{\PlotData}[2]{%
\begin{tikzpicture}%
\draw (0cm, 0cm) node [anchor=south west] {\fbox{\includegraphics[width=\PicWidth]{#2}}};%
\draw (0cm, 0cm) node [anchor=south west] {\fbox{\includegraphics[width=0.75cm]{#1}}};%
\end{tikzpicture}}%
\begin{figure}[t]%
\centering%
\ifexternalized%
    \includegraphics{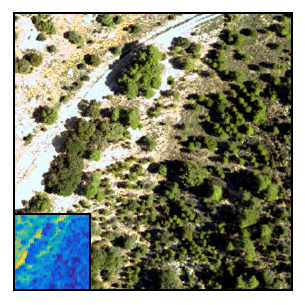}\hspace{\PicDistance}%
    \includegraphics{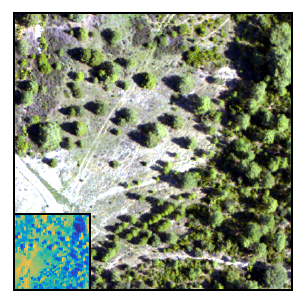}\hspace{\PicDistance}%
    \includegraphics{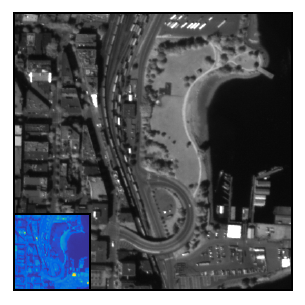}\\[-3mm]%
\fi%
\caption{Three example data for image fusion in remote sensing. They each consist of a hyperspectral image (small image, only one channel shown) and an image of higher spatial resolution (large image). The goal is to create an image that has both high spatial and high spectral resolution.}\label{FIG:GOAL}%
\end{figure}}%

Most image fusion techniques are based on certain assumptions regarding the data and the corresponding imaging devices. First, it is often assumed that the image of high spatial resolution is a linear combination of the spectral channels with known weights \cite{Loncan2015}. Second, the loss of resolution is usually modeled as a linear operator which consists of a subsampled convolution with known kernel (point spread function). While both assumptions may be justified in some applications, it may be difficult to measure or estimate the weights and the convolution kernel in a practical situation. 

Instead of having accurate descriptions or measurements about the system response of our cameras \cite{Ballester2006, Loncan2015, Hou2016}, we make use of the fact that both images are acquired from the same scenery. While the actual image intensities will depend on the wavelengths that are being recorded and are therefore dissimilar, the geometric structure of those images is likely to be very similar as they are both taken from the same terrain \cite{Ballester2006, Moeller2009, moller2012variational, Hou2016}. This approach has been studied extensively in the context of medical imaging where traditionally an anatomical imaging modality with high spatial resolution is used to aid the reconstruction of a functional imaging modality, cf. e.g. \cite{Kaipio1999, Bowsher2004, Vunckx2012, Ehrhardt2016pet, Ehrhardt2016mri, Kolehmainen2017, Mehranian2017}. 

\begin{figure}%
\centering%
\ifexternalized%
    \includegraphics{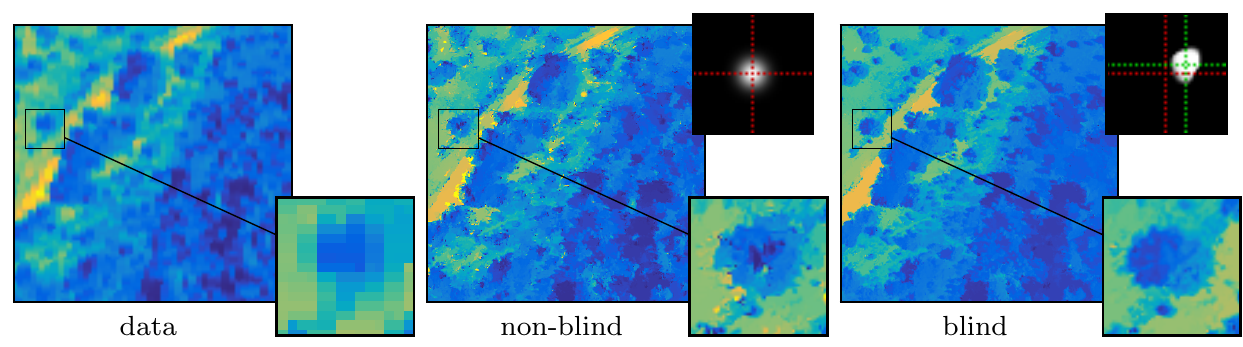}\\[-3mm]%
\fi
\caption{\rev{Comparison of \emph{non-blind} versus \emph{blind} image fusion. \textbf{From left to right:} hyperspectral image (one channel, four times enlarged), result of non-blind and blind image fusion together with the kernels that are used or estimated, respectively. The plotted kernels are of size $41\times 41$ and have been enlarged for better visibility. The dotted green lines highlight the center of mass of the obtained kernel and indicate the amount of translational displacement between the fused images.}} \label{FIG:blind_vs_non-blind}%
\end{figure}

Note that the hyperspectral data image and the high-resolution photograph usually undergo a georectification procedure which aligns them using a digital elevation map of the study area. However, this adjustment is often imprecise, particularly in topographically complex regions, so further steps are required to co-register the images precisely \cite{Lee2015}. Moreover, having a good description of the point spread function of the system is very unnatural. Thus, we take a blind approach and estimate the point spread function within our reconstruction and thereby learn this information from the recorded data. This approach is intrinsically related to blind deconvolution \cite{Amizic2013, Perrone2016, Kundur1996blind, Vorontsov2017blind}. \Fref{FIG:blind_vs_non-blind} shows an example of blind vs non-blind image fusion applied to the real remote sensing data with unknown blurring kernel. It can be seen that the shape of the kernel that is estimated during the reconstruction clearly differs from the fixed kernel. Equally important is the translation of the estimated kernel, which can compensate for shifts between the given image pair. To the best of our knowledge this is the first contribution where structural side information / anatomical priors are combined with registration which removes the unnatural assumption of perfectly registered images that is usually made, cf. \cite{Kaipio1999, Bowsher2004, Vunckx2012, Ehrhardt2016pet, Ehrhardt2016mri, Kolehmainen2017, Mehranian2017}, for instance.

\section{Mathematical Model and Notation}
In this work we consider the problem of simultaneous image fusion and blind image deconvolution which can be cast as the optimization problem
\begin{equation}\label{EQ:BID}
(u^\ast, k^\ast) \in \Argmin{(u, k)\in \spaceLarge\times\spaceKernel}{\left\{\frac 12 \|\mA_k u - f\|^2 + \Ru(u) + \Rk(k)\right\}}
\end{equation}
where the \emph{forward operator} $\mA_k$ is a sampled blur with kernel $k$. Thus, it relates an image $u \in \spaceLarge := \R^m$ of size $m = (m_1, m_2)$ to blurred and subsampled data $f \in \R^n, n = (n_1, n_2)$. Note that we used \emph{multi-index notation} which means $\R^m = \R^{m_1 \times m_2}$, for instance. Here, we consider the case of \emph{super-resolution} $n < m$ (true if and only if $n_1 < m_1$ and $n_2 < m_2$). The data discrepancy is measured in terms of the Euclidean / Frobenius norm $\|x\|^2 := \sum_{i} x_i^2$ and the optimal solution is \emph{regularized} through the penalty functionals $\Ru$ and $\Rk$.

Since---up to now---the ingredients of our model are only vaguely defined, the remainder of this section is devoted to an detailed explanation of the operators and regularization functionals involved.

\subsection{The Forward Operator}
\label{SEC:forward_op}
The forward operator $\mA_k \De \mS \circ \mB \circ \mC_{k}$ is the composition of three operators: $\mC_k$ represents a \emph{convolution} with a kernel $k$, $\mB$ performs \emph{boundary treatment} and $\mS$ is a \emph{sampling operator}. 

The convolution operator $\mC_k: \spaceLarge \to \spaceLarge$ with kernel $k \in \spaceKernel := \R^r, r := (r_1, r_2)$ is defined by 
\begin{equation}
\mC_k u \De \mJ k \ast u 
\end{equation}
where $\ast$ is the cyclic convolution of images in $\spaceLarge = \R^m$. Since the kernel $k$ is generally assumed to have smaller support than the image ${u}$, i.e. $r < m$, we introduced the embedding operator $\mJ : \spaceKernel \rightarrow \spaceLarge$ which embeds the \enquote{small} convolution kernel $k \in \spaceKernel$ into the image space $\spaceLarge$ by padding it with zeros. By assuming periodic boundary conditions on $u$ and $\mJ k$, the convolution can be efficiently computed via the discrete Fourier transform $\Fourier$, i.e. $\mC_k u = \Fourier^{-1}\left(\Fourier \mJ k \odot \Fourier u\right)$ where $\odot$ denotes the Hadamard product (pointwise multiplication). With a slight abuse of notation, we can exploit the symmetry of the convolution to infer
\begin{equation}
\mC_k u = \mJ k \ast u = u \ast \mJ k = \mC_u \mJ k \, ,
\end{equation}
and thus
\begin{equation}\label{EQ:symmetry}
\mA_k u = \mA_u \mJ k \, ,
\end{equation}
which will prove helpful later-on.

While periodic boundary conditions are computationally useful, any kind of boundary treatment is artificial and may result in artifacts. For instance, boundary artifacts are expected in the boundary region with margins $l \De (r-1)/2$ where $r$ is the diameter of the convolution kernel. To avoid these artefacts, we follow the \emph{boundary-layer approach} (cf. \cite{chan2005image} and references therein; also described by Almeida and Figueiredo in \cite{almeida2013deconvolve}) and introduce a boundary clipping (margin deletion) operator $\mB:\spaceLarge \to \R^{m-2l}$, 
\begin{equation}
(\mB u)_i \De u_{i + l} 
\end{equation}
that maps a large image $u$ to a \emph{meaningful image} $\mB u \in \R^{m-2l}$ which is independent of the type of boundary conditions used for the precedent convolution. Thus, after computation of an optimal solution $u^\ast$ of \eref{EQ:BID} we only use the meaningful part $\mB u^\ast$.

Finally, the third component of our forward operator is the sampling $\mS:\R^{m-2l}\to\R^n$ which mimics an integration sensor of size $s \times s$, i.e. 
\begin{equation}
    (\mS u)_i \De \frac{1}{s^2}\sum_{j \in \Delta_i} u_j
\end{equation}
where the detector indices are given by the set $\Delta_i := \{ j = (j_1, j_2) \, | \, i \leq j < i + s \}$, which contains $s^2$ elements. Consequently, for a kernel of odd dimensions the relation $m - 2l = s n$ connects the sizes of kernel, image and data.

Let us now turn to the regularization and prior knowledge on the image $u$ and the kernel $k$.

\subsection{Regularization Functionals} \label{SEC:regularisers} 
\subsubsection{Image Regularization} A popular prior to regularize images is the \emph{total variation} \cite{Rudin1992ROF} as it allows the recovery of smooth images whilst preserving edges. It is defined as the global 1-norm of the discretized gradient
\begin{equation}\label{EQ:TV}
\tv(u) \De \sum_i\|\grad u_i\| \, .
\end{equation}
Due to the periodic boundary conditions needed for the forward operator, the discretized gradient (with forward differences) $\grad u \in \spaceLarge^2$ of an image $u \in \spaceLarge$  is
\begin{equation}
(\grad u_i)_j \De u_{\operatorname{mod}(i+e_j, m)}-u_i \, ,\quad j\in\{1,2\},
\end{equation}
where $e_1 = (1,0)$ and $e_2 = (0,1)$ are the standard unit vectors in $\R^2$.

\begin{figure}%
\centering%
\ifexternalized%
    \includegraphics{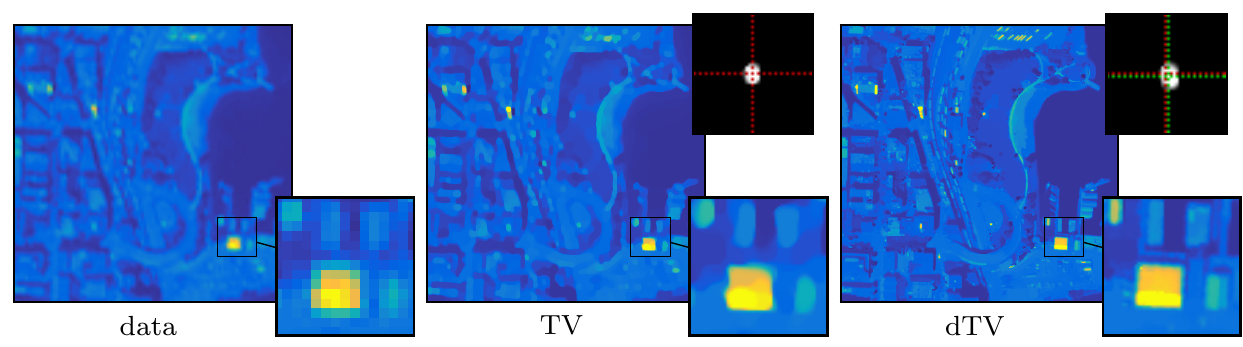}\\[-3mm]
\fi
\caption{Comparison of total variation ($\tv$) versus directional total variation ($\dtv$) regularization for image fusion. \textbf{From left to right:} hyperspectral image (one channel, four times enlarged), results of $\tv$ and $\dtv$ together with the estimated kernels.} \label{FIG:TV_vs_dTV}
\end{figure}

In the task at hand, the high-resolution photograph can be viewed as a source of additional a-priori knowledge about the sought solution. As both the hyperspectral data and the high-resolution photograph depict the same scenery, it is reasonable to assume that they show the similar geometrical structures albeit having very different intensities. However, the TV functional \eref{EQ:TV} is not able to incorporate this kind of additional prior information. Instead, we will apply the so-called directional TV (dTV), which can fuse the structural information of a high-resolution image with the intensities of a low-resolution channel of a hyperspectral image. Figure~\ref{FIG:TV_vs_dTV} provides a visual comparison of reconstructions obtained from applying the TV and the dTV functional, respectively.

\begin{definition*}[Directional Total Variation \cite{Ehrhardt2016mri}]\label{DEF:dTV}
Let $\xi \in \spaceLarge^2, \|\xi_i\| \leq \gamma < 1$ be a vector-field that determines directions and denote by $\mP \in \spaceLarge^{2 \times 2}, \mP_i := \mI-\xi_i \otimes \xi_i $ an associated matrix-field where $\mI$ is the $2\times 2$ identity matrix and $\otimes$ marks the outer product of vectors. Then the \emph{directional total variation} $\dtv : \spaceLarge \to \R$ is defined as as
\begin{equation}\label{EQ:dTV}
\dtv(u) \De \sum_i \|\mP_i \grad u_i \| \, .
\end{equation}
\end{definition*}
Note that due to the linearity of $\mP_i$ the directional total variation is \emph{convex}.
Let us shortly comment on the interpretation of the directional total variation. The vector-field $\xi$ allows the definition of directions that should be less penalized and therefore get promoted. It is easy to see that the quantity $\mP_i \grad u_i$ can be expanded as
\begin{equation}
\mP_i \grad u_i = \grad u_i - \langle \xi_i, \grad u_i \rangle \xi_i
\end{equation}
where $\langle \cdot, \cdot \rangle$ denotes the inner product on $\R^2$. This expression reduces to $(1-\|\xi_i\|^2)\grad u_i$ in regions where $\grad u_i$ is collinear to $\xi_i$, and to $\grad u_i$ where $\grad u_i$ is orthogonal to $\xi_i$. Thus, gradients that are aligned / collinear $\xi_i$ are favored as long as $\|\xi_i\| > 0$. In the extreme case $\|\xi_i\| = 1$, aligned gradients are not penalized any more. It is important to note, that this prior \emph{does not enforce} gradients in the direction of $\xi_i$, as a gradient in the direction of $\xi_i$ is never \enquote{cheaper} than a zero gradient. 

The uniform upper bound $\gamma < 1$ ensures that the directional total variation is equivalent to the total variation in a semi-norm sense, i.e.
\begin{equation}
    (1-\gamma^2) \tv(u) \leq \dtv(u) \leq \tv(u) \, .
\end{equation}

Similar versions of the directional total variation have been used before \cite{Grasmair2010,Ehrhardt2016pet, Kolehmainen2017} and it is related to other anisotropic priors \cite{Kaipio1999,Estellers2013, Estellers2015} and the notion of parallel level sets \cite{Ehrhardt2014c, Ehrhardt2015, Ehrhardt2015PhD}. Note that the directional total variation generalizes the usual total variation \eref{EQ:TV} for $\xi = 0$ and other versions of the directional total variation \cite{Bayram2012,Kongskov2017} where the direction $\xi$ is constant and not depending in the pixel location $i$. With the isotropic choice $\mP_i = w_i \mI,\, 0 \leq w_i \leq 1$ it reduces to weighted total variation \cite{Arridge2008,Grasmair2009,Knoll2010,Grasmair2010,Ehrhardt2016mri}.

{%
\newcommand{\PlotSideInfo}[1]{%
\def\ZoomWidth{2.5cm}
\begin{tikzpicture}[spy using outlines]
\draw (0cm, 0cm) node [anchor=south west] {\fbox{\includegraphics[clip, trim=5cm 9.5cm 4cm 8.5cm, width=\PicWidth]{#1}}};%
\spy [black, draw, width=2.5cm, height=2cm, magnification = 3, connect spies] on (2.3,2.1) in node [left] at (2.2,0.3);
\end{tikzpicture}}%
\begin{figure}%
\centering%
\ifexternalized%
    \includegraphics{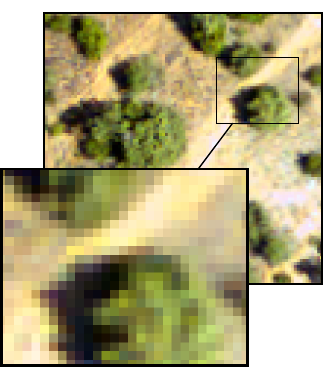}\hspace{\PicDistance}%
    \includegraphics{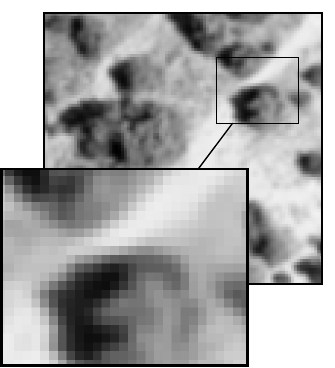}\hspace{\PicDistance}%
    \includegraphics{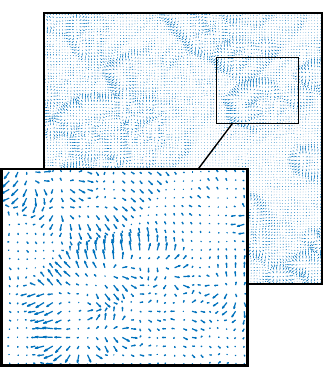}\\[-3mm]%
\fi
\caption{Side information for Directional Total Variation. The side information is given as an RGB image \textbf{(left)} and converted into gray scales \textbf{(center)} from which a vector-field may be computed \textbf{(right)}, e.g. with \eref{EQ:vectorfield} (shown for $\gamma = 1, \varepsilon = 0.3$). The close-ups show that the vector-field does not only carry detailed information about edge \emph{locations} but also about edge \emph{directions}.}\label{FIG:vectorfield}%
\end{figure}%
}%
Having this definition of the directional total variation at hand, the question is how to define the vector-field $\xi \in \spaceLarge^2$. As we want to favor images having a similar structure as the high-resolution photograph, we let $v\in\spaceLarge$ denote a gray-scale image which is generated from the color high-resolution image and define the vector-field
\begin{equation}\label{EQ:vectorfield}
\xi_i \De \gamma\frac{\grad v_i}{\|{\grad v}_i\|_\varepsilon} \, ,
\end{equation}
where $\|x\|_\varepsilon \De \sqrt{\|x\|^2+\varepsilon^2}$ and the scaling $\gamma \in [0, 1)$ ensures that $\|\xi_i\| \leq \gamma$. The parameter $\varepsilon > 0$ guarantees that the vector-field $\xi$ is well-defined, even if $\grad v_i = 0$. Moreover, it allows to control which gradient magnitudes $\|\grad v_i\|$ should be considered meaningful and which are merely due to noise. Thus, both parameters $\varepsilon$ and $\gamma$ in general will depend on the given side information $v$ but can be chosen without considering a particular data set $f$. \rev{Note that in practice, one chooses $\gamma$ very close or even equal to 1 (cf. \cite{Ehrhardt2016mri} for the latter) such that the side information image has sufficiently much influence. Extensive studies of this parameter are performed in \cite{Kongskov2017} where the authors obtain their best results for small values of $a\De1-\gamma^2$.} An example photograph, its gray scale version and the corresponding vector-field are shown in \fref{FIG:vectorfield}.  

In addition to the structural a-priori knowledge given by the photograph and encoded in the directional total variation, it is reasonable to assume that the solution we are looking for is non-negative. Thus, the regularization functional for the image becomes 
\begin{equation}\label{EQ:Reg_u}
    \Ru \De \RegParam_u \dtv + \imath_{[0,\infty)^m}
\end{equation}
where the characteristic function of the non-negative quadrant is defined as
\begin{equation}\label{EQ:char_pos_quadr}
    \imath_{[0,\infty)^m}(u) \De \begin{cases*} 0 & if $u \in [0,\infty)^m$ \\ \infty & else \end{cases*} \, .
\end{equation}
The regularization parameter for the image $\RegParam_u>0$ controls the trade-off between regularity and data fidelity.

\subsection{Kernel Regularization}
Similarly as for the image $u$, it is reasonable to assume that the kernel is non-negative and regular. Motion blurs that arise from ideal diffraction-free lenses have typically a compactly supported kernel with sharp cut-off boundaries \cite[p. 234]{chan2005image}. Thus, the desired regularity of the kernel may be achieved by utilizing the total variation. Of course, other smoothness priors such as the $H^1$-semi-norm or total generalized variation \cite{Bredies2010} are possible, too, and will be subject of future work. In addition, blind deconvolution has the intrinsic non-uniqueness in the problem formulation that does not allow to estimate the correct global scaling factors for both the kernel and the image. To circumvent this issue, a common choice (e.g. \cite{Kundur1996blind,Pock2016iPALM,Vorontsov2017blind,chan2005image}) is to normalize the kernel with respect to the 1-norm. Thus, together with the non-negativity, we assume that the kernel is in the unit simplex
\begin{equation}
\Simplex \De \left\{ k \in \spaceKernel \, : \, k_i \geq 0, \, \sum_i k_i = 1\right\} \, .
\end{equation}
Letting $\imath_{\Simplex}$ denote the characteristic function of the unit simplex, defined analogously to \eref{EQ:char_pos_quadr}, the regularization functional for the kernel becomes
\begin{equation}\label{EQ:Reg_k}
    \Rk \De \RegParam_k \tv + \imath_{\Simplex}
\end{equation}
where we again can trade-off regularity and data fidelity by the regularization parameter $\RegParam_k>0$ which in general needs to be chosen differently than the regularization parameter for the image $\RegParam_u$.

\section{Algorithms}
Given the abstract mathematical model \eref{EQ:BID} which models the solution that we are after, this section discusses algorithms that will attempt to numerically compute this solution. For a concise presentation, we cast problem \eref{EQ:BID} in the form
\begin{equation}\label{EQ:BID_short}
(u^\ast, k^\ast) \in \Argmin{u, k}{\Obj(u,k)}
\end{equation}
by defining $\Obj(u,k)\De \DataFid(u,k) + \Ru(u)+\Rk(k)$ and abbreviating the data fidelity term in \eref{EQ:BID} by $\DataFid(u,k)\De\frac{1}{2}\|\mA_ku-f\|^2$. While the function $\Obj$ is not convex in the joint variable $(u,k)$, it is bi-convex, i.e. it is convex in each of the variables $u$ and $k$. The latter holds true as the regularization functions $\Ru$ and $\Rk$ are convex and the data fidelity $\DataFid$ is bi-convex. 

An abstract algorithm for finding solutions of \eref{EQ:BID_short} is the \emph{proximal alternating minimization (PAM)} algorithm \cite{Attouch2010PAM}. Given an image-kernel pair $(u, k)$ and step sizes $\tau_u, \tau_k > 0$, it computes the next iterate $(u^+, k^+)$ by the update scheme
\begin{subequations}\label{EQ:PAM}
\begin{eqnarray}
u^+ = \Argmin{x}{\left\lbrace \frac12 \|x - u\|^2 + \tau_u \Obj(x, k) \right\rbrace} \, , \\
k^+ = \Argmin{x}{\left\lbrace \frac12 \|x - k\|^2 + \tau_k \Obj(u^+, x) \right\rbrace} \, .
\end{eqnarray}
\end{subequations}
Under reasonably mild conditions on the regularity of $\Obj$ (see \cite{Attouch2010PAM} for more details), PAM monotonously decreases the objective and its iterates converge to a critical point of $\Obj$. As the overall problem is non-convex, we cannot hope to find global solutions with this strategy.

\subsection{Proximal alternating linearized minimization (PALM)}
In addition, we may exploit the fact that the data term $\DataFid$ is continuously differentiable with respect to $u$ and $k$ and that it is not very difficult to compute the \emph{proximal operators} with respect to the regularization functionals $\Ru$ and $\Rk$. The proximal operator of a functional $\mathcal R$ is defined as the solution operator to a denoising problem with regularization term $\mathcal R$
\begin{equation}\label{EQ:prox}
\Prox{\RegFun}(y)\De \Argmin{x}{\left\lbrace \frac12\|x-y\|^2 + \RegFun(x)\right\rbrace} \, .
\end{equation}
These extra information can be utilized with the \emph{proximal alternating linearized minimization (PALM)} \cite{Bolte2014PALM}. Linearizing the differentiable part $\DataFid$ of the objective function $\Obj$ in the update of \eref{EQ:PAM} yields the new update
\begin{subequations}
\begin{eqnarray}\label{EQ:PALM1}
u^+ = \Prox{\stepsize_u \Ru}\Bigl(u - \stepsize_u \grad_u \DataFid(u, k) \Bigr) \,, \label{EQ:PALM2} \\
k^+ = \Prox{\stepsize_k \Rk}\Bigl(k - \stepsize_k \grad_k \DataFid(u^+, k) \Bigr) \, .
\end{eqnarray}
\end{subequations}
It can be interpreted both as a linearized version of PAM and as an alternating form of forward-backward splitting \cite{Lions1979,combettes2005signal}. Note that PALM also monotonously decreases the objective and has the same convergence guarantees as long as the step size parameters are well-chosen which we will discuss later in this section.

\subsection{Inertial PALM (iPALM)}
Local methods for non-convex optimization may lead to critical points that are not global minimizers. The authors of \cite{Ochs2014iPIANO} proposed an inertial variant of forward-backward splitting which next to the gradient direction gives inertia to steer the iterates in the previously chosen direction. They empirically showed on a simple two dimensional problem that inertia may help to avoid spurious critical points. An inertial version of PALM \cite{Pock2016iPALM}, called \emph{iPALM}, follows the updates
\begin{subequations}
\begin{eqnarray}
u_\InertParamA &= u + \InertParamA(u - u^-) \, , \\
u^+ &= \Prox{\stepsize_u \Ru} \Bigl(u_\InertParamA - \tau_u \grad_u \DataFid(u_\InertParamA, k)\Bigr) \, ,\\
k_\InertParamA &= k + \InertParamA(k - k^-) \, ,\\
k^+ &= \Prox{\stepsize_k \Rk} \Bigl(k_\InertParamA - \tau_k \grad_k \DataFid(u^+, k_\InertParamA)\Bigr) \, ,
\end{eqnarray}
\end{subequations}
with inertia parameter $\InertParamA \in [0,1)$. Next to the current iterate $(u, k)$ it also depends on the previous iterate $(u^-, k^-)$. We remark that PALM can be recovered by vanishing inertial parameter $\InertParamA = 0$. Also note that one may define iPALM not with one inertial parameter $\alpha$ but with four different ones \cite{Pock2016iPALM}. While iPALM is \emph{not} guaranteed to decrease the objective $\Obj$, it monotonously decreases a surrogate functional that relates to $\Obj$ \cite{Pock2016iPALM} and its iterates are guaranteed to converge to a critical point of $\Obj$ for well-chosen step size and inertia parameter. 

\begin{algorithm}%
\caption{\textbf{PALM} / \textbf{iPALM} with backtracking. PALM is recovered by setting the inertial parameter $\InertParamA = 0$. The parameters $\underline{L}, \overline{L}, \eta, \theta$ can all be chosen different for the image $u$ and kernel $k$ but for simplicity we chose not to.}\label{ALG:iPALM}%
\begin{algorithmic}[1]%
\footnotesize
\Require initial iterates $u, k$, inertial parameter $\InertParamA \in [0,1)$, initial step size parameters $L_u,L_k$, constant for step size $\theta > 1$, bounds for step size parameters $0 < \underline{L} \leq \overline{L} < \infty$, backtracking constant $\eta> 1$
\Function{iPALM}{}
    \State $u^- \gets u,\;k^-\gets k$
    \For{$\iter=0, 1, \ldots$}
    \State {\color{gray}\textit{\% update image $u$}}
    \State $ u_\InertParamA \gets u + \InertParamA(u - u^-)$ 
    \State $(u^+, L_u) \gets$ \Call{backtracking}{$u_\InertParamA, \grad_u \DataFid(u_\InertParamA, k), L_u$}
    \State $u^-\gets u,\;u\gets u^+$
    \State {\color{gray}\textit{\% update kernel $k$}}
    \State $ k_\InertParamA \gets k + \InertParamA(k - k^-)$ 
    \State $(k^+, L_k) \gets$ \Call{backtracking}{$k_\InertParamA, \grad_k \DataFid(u^+, k_\InertParamA), L_k$}
    \State $k^-\gets k,\;k\gets k^+$
    \EndFor
\State \textbf{return} $(u^+,k^+)$
\EndFunction

\Function{backtracking}{$x_\alpha, g, L$}
    \While{True}
    \State {\color{gray}\textit{\% select step size}}
    \State $\stepsize \gets \frac{1-\InertParamA}{1 + 2\InertParamA}\frac{2}{\theta L}$
    \State {\color{gray}\textit{\% forward-backward step where the proximal operator}}
    \State {\color{gray}\textit{\% is approximated by a fixed number of warm-started iterations}}
    \State $x^+ \gets \Prox{\stepsize \RegFun}(x_\alpha - \stepsize g)$
    \If{Descent inequality~\eref{EQ:BT:DESCENTLEMMA} does \textbf{not} hold}
        \State {\color{gray}\textit{\% increase Lipschitz constant and try again}}
        \State $L \gets \min(\eta L, \overline{L})$
        \State \textbf{continue}
    \EndIf
    \If{Proximal descent inequality~\eref{EQ:BT:PROX} does hold}
        \State {\color{gray}\textit{\% continue to next iteration with larger step size}}
        \State $L \gets \max(L/\eta, \underline{L})$
        \State \textbf{break}
    \EndIf
    \EndWhile
\State \textbf{return} $(x^+, L)$
\EndFunction%
\end{algorithmic}%
\end{algorithm}%
\subsection{Step sizes and backtracking}
The convergence of PALM and iPALM depend on the step size parameters $\tau_u, \tau_k$. Following \cite{Pock2016iPALM}, we choose the step size parameters as
\begin{equation}\label{EQ:stepsizes}
\stepsize_x \De  \frac{1-\InertParamA}{1 + 2\InertParamA}\frac{2}{\theta L_x} \,, \qquad x \in \{u, k\} \, ,
\end{equation}
where $\theta>1$ is a global constant and $L_x$ is the local Lipschitz constant of the derivative of the data fit $\mathcal D$ with respect to $x$. Using the relation \eref{EQ:symmetry}, its derivatives with respect to the image and kernel are
\begin{subequations}\label{EQ:gradD}
\begin{eqnarray}
\nabla_u \DataFid(u,k) &= \mA_k^\ast(\mA_ku-f) \, , \\
\nabla_k \DataFid(u,k) &= \mJ^\ast\mA_u^\ast(\mA_u\mJ k-f) \,
\end{eqnarray}
\end{subequations}
which shows that $L_u$ will depend on $k$ and $L_k$ on $u$. While it is possible to conservatively estimate these constants, standard estimates turn out to be highly pessimistic and one can do significantly better by determining $L_u$ and $L_k$ with a backtracking scheme \cite{Ochs2014iPIANO,Beck2009fista}. The actual local Lipschitz constants $L_u$ and $L_k$ satisfy the descent inequalities (also called descent Lemma) \cite{Beck2009fista, Ochs2014iPIANO, Pock2016iPALM, Bolte2014PALM}
\begin{subequations} \label{EQ:BT:DESCENTLEMMA}
\rev{
\begin{eqnarray}
\fl \DataFid(u^+, k) \leq \DataFid(u_\alpha, k) + \left\langle\grad_u \DataFid(u_\alpha, k), u^+ - u_\alpha\right\rangle + \frac{L_u}{2}\|u^+-u_\alpha\|^2 \, , \\
\fl \DataFid(u^+, k^+) \leq \DataFid(u^+, k_\alpha) + \left\langle\grad_k \DataFid(u^+, k_\alpha), k^+ - k_\alpha\right\rangle + \frac{L_k}{2}\|k^+-k_\alpha\|^2 \, .
\end{eqnarray}}
\end{subequations}
If during the iterations these inequalities are not satisfied, we increase the Lipschitz estimates by a constant factor $\eta > 1$ and repeat the iteration.

In addition to the descent inequality, an inexact evaluation of the proximal operators (e.g. due to finite iterations) may also hinder convergence. Thus, we will evaluate the proximal operators to a precision such that the proximal descent inequalities
\begin{subequations} \label{EQ:BT:PROX}
\begin{eqnarray}
\fl \RegFun_u(u^+)&\leq \RegFun_u(u) + \left\langle \grad_u\DataFid (u_\InertParamA, k), u - u^+\right\rangle + \frac{1}{2 \stepsize_u}\left(\|u - u_\InertParamA\|^2 - \|u^+-u_\InertParamA\|^2\right) \, , \\
\fl \RegFun_k(k^+)&\leq \RegFun_k(k) + \left\langle \grad_k\DataFid (u^+,k_\InertParamA), k - k^+\right\rangle + \frac{1}{2 \stepsize_k}\left(\|k - k_\InertParamA\|^2 - \|k^+-k_\InertParamA\|^2\right)
\end{eqnarray}
\end{subequations}
hold. Equations \eref{EQ:BT:DESCENTLEMMA}, \eref{EQ:BT:PROX} and the parameters $L_u, L_k$ guarantee the monotonic descent of PALM and iPALM with respect to the objective or its surrogate \cite{Pock2016iPALM, Bolte2014PALM}.

The pseudo code for PALM and iPALM with this backtracking can be found in algorithm \ref{ALG:iPALM}. 

\section{Numerical Setting}
\paragraph{Data} In this work we use data from two separate data sets. The first data set \cite{NERC2014} is from environmental remote sensing to study vegetation in the Mediterranean woodlands in Alta Tajo natural park, northeast of Madrid, Spain \cite{Simonson2016}. It was acquired in October to November 2014 using an AISA Eagle hyperspectral sensor and a Leica RCD105 39 megapixel digital camera mounted on a Dornier 228 aircraft. The hyperspectral imagery consisting of 126 spectral bands almost equidistantly covering the wavelengths 400 nm to 970 nm with a spatial resolution of 1 m x 1 m. The aerial photograph has a resolution of 0.25 m x 0.25 m but only captures channels corresponding to the visible red, green and blue. We refer to images from this data set as \DataTreesA, \DataTreesB.

The second data set \cite{GRSS2016, Mou2017} shows an urban area of Vancouver, Canada. It was acquired in March to May 2015 by the DEIMOS-2 satellite operating on a mean altitude of 620 km. The satellite carried a very highly resolving push-broom camera with one panchromatic channel of 1 m x 1 m resolution and the four multispectral channels red, green, blue and near-infrared with a resolution of 4 m x 4 m. Images from this data set are marked \DataUrbanA, \DataUrbanB.

Next to the already described real data sets we simulated an environmental data set where we know the ground truth image and kernel. We test two different kernels, one which is disk-shaped and one which is an off-centered Gaussian and refer to the corresponding data sets as \DataGTdisk~and \DataGTgaussian. The red channel of an RGB image is convolved with these two kernels, subsampled by a factor of $s=4$ and Gaussian noise of variance 0.001 is added. The data is shown at the top left of figures \ref{FIG:ground_truth_results_disk} and \ref{FIG:ground_truth_results_gauss}. For the data set \DataGTgaussian~we use the RGB image itself as side information and for \DataGTdisk~we use a shifted version of it. This situation occurs frequently in applications since usually data and side information are acquired by different imaging devices which can lead to a relative translation between the observed images. Note that we used replicated instead of periodic boundary conditions for the convolution in order to avoid \emph{inverse crime}. As the ground truth is known in these cases, we evaluate the result in terms of the \emph{structural similarity index (SSIM)} \cite{Wang2004} and the \emph{Haar wavelet-based perceptual similarity index (HPSI)} \cite{reisenhofer2016haar}. \rev{As compared to measures like the peak-signal-to-noise ratio (PSNR) or the mean squared error (MSE), the SSIM and the HPSI aim at reproducing how image similarity is perceived by human viewers. The SSIM is one of the most widely used image similarity measures and obtained by comparing local statistical parameters such as local means and local variances. The HPSI is based on a simple discrete Haar wavelet transform and incorporates basic assumptions about the human visual system. The HPSI was recently shown to yield state-of-the-art correlations with human opinion scores on large benchmarking databases of differently distorted images.}

In all numerical experiments the data $f$ are scaled to $[0,1]$.

\paragraph{Visualization} Throughout the results, the data and reconstructed images are shown with MATLAB's \enquote{parula} colormap where yellow represents values $\geq 1$ and dark blue corresponds to values $\leq 0$, respectively. The kernels are visualized with gray scales where black corresponds to the smallest value and white to the largest. In addition, we draw a red hair cross on the center, i.e. the pixel with indices $l+1$, where $l$ denotes the radius of the kernel. The kernel's centroid $\sum_i i k_i$ is visualized by a green hair cross in order to visualize the off-set between data and side information.

\paragraph{Algorithms} The numerical results will be computed with the algorithms PALM, iPALM and PAM. For PAM, the inner problems require solving a deconvolution problem with convex regularization which is implemented with ADMM \cite{Gabay1976,Boyd2010,Afonso2010} following a similar strategy as in \cite{Ehrhardt2016mri}. All three---PALM, iPALM and PAM---require the computation of proximal operators for the total variation and directional total variation under a convex constraint. These are implemented with a finite number of warm started Fast Gradient Projection/FISTA \cite{Beck2009fistaSIAM,Beck2009fista} iterations. Within these, projections onto the constraint sets are computed by $\Prox{\imath_{[0,\infty)^m}}(u) = \max(0,u)$ (to be understood componentwise) and the fast algorithm of \cite{Duchi2008efficient} for the projection onto the unit simplex. A MATLAB implementation of the algorithms and the data itself can be found at \cite{Supp2017}.

\paragraph{Parameters}
The numerical experiments require model parameters and algorithm parameters. For the mathematical model, the parameters of the directional total variation are chosen as $\gamma = 0.9995$ and $\varepsilon = 0.003$ for all experiments, cf.~\eref{EQ:vectorfield} and \fref{FIG:GAMMA} in the appendix. The low resolution images are of size $100\times 100$ with an subsampling rate of $s=4$. Since we fix the kernel size to be $41\times 41$, we will consequently obtain reconstructions of size $440\times 440$ where we only visualize the meaningful part of size $400\times 400$. 

Except for the algorithm comparison, we use PALM with 2000 iterations and a step size constant $\theta = 1.1$, cf.~\eref{EQ:stepsizes}. The backtracking parameters in algorithm~\ref{ALG:iPALM} are chosen as $\eta=2$, $\underline{L} = 1$ and $\overline{L} = 10^{30}$.

\paragraph{Initialization} 
Both algorithm \ref{ALG:iPALM} and our implementation of PAM require initial guesses for image and kernel. \rev{While the kernel is initialized with a compactly supported Gaussian kernel, we choose the initial image to be $\mH f$, where $\mH:\R^n\to\spaceLarge$ is a right inverse of the operator $\mS\circ\mB$, i.e $\mH$ is an upsampling operator (cf. the top-left images in figures \ref{FIG:ALGCOMP:GT} and \ref{FIG:ALGCOMP:TREES}).} 
\section{Numerical Results}
\subsection{Comparison of Algorithms}
{%
\newcommand{\PlotAlg}[6]{
\PlotKernelSpy{#2}{\FilenameLocal__image_#1_parula.png}{\FilenameLocal__kernel_#1.png}{#5}{#6}{\SpyX}{\SpyY}%
\draw (#5+.1cm, #6-.5cm) node [anchor=west] {\footnotesize $\Obj=$ #3, \footnotesize SSIM: #4};}%
\begin{figure}[h]%
\def\SpyX{2.4cm}\def\SpyY{0.8cm}
\def\FilenameLocal{trees1_shift_5px_disk__PALM0__lambda_u__0-1__lambda_k_10__gamma_0-9995__eta_0-003}
\centering%
\ifexternalized%
    \includegraphics{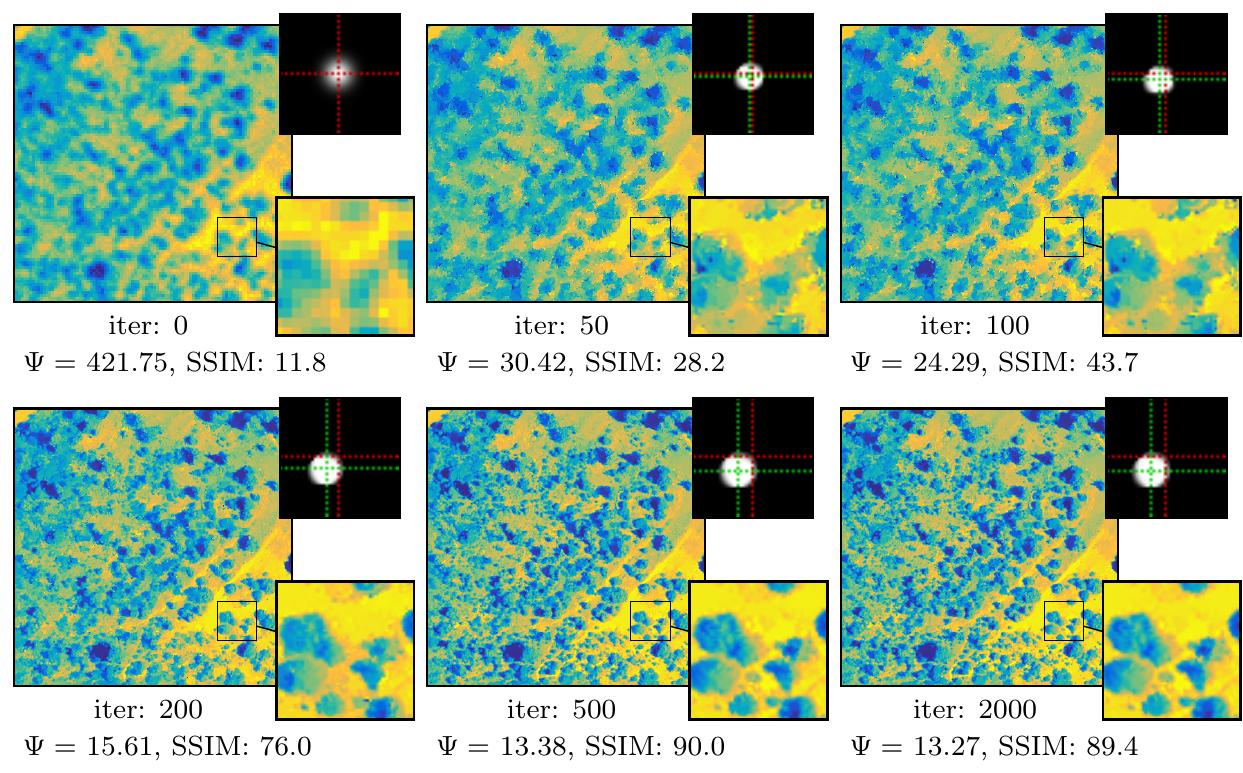}\\[-3mm]
\fi
\caption{Iterates of PALM for data set \DataGTdisk~with $\lambda_u = 0.1$, $\lambda_k = 10$ with objective function values $\Obj$ and similarity to ground truth.} \label{FIG:ALGCOMP:GT}%
\vspace*{3mm}
\ifexternalized%
    \includegraphics{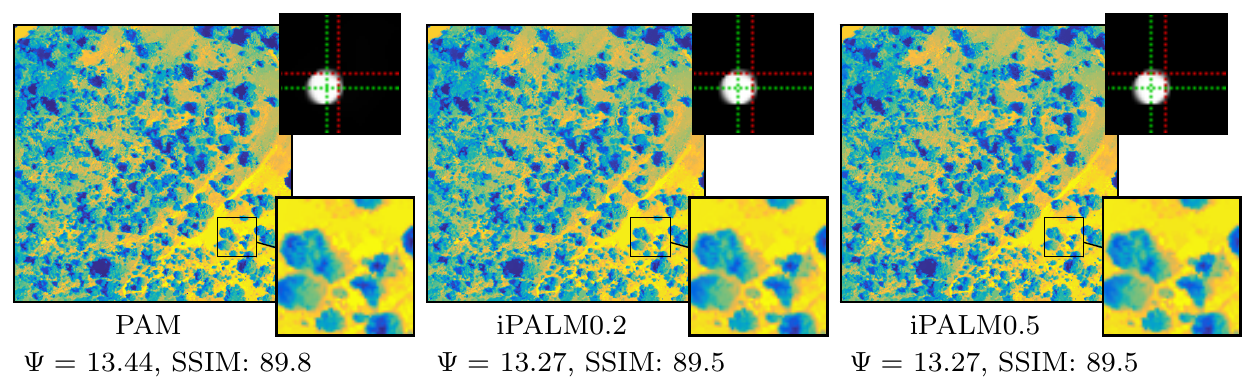}\\[-3mm]
\fi
\caption{Reconstructions by three different algorithms---PAM and iPALM with inertial parameters $\alpha = 0.2$ and $\alpha = 0.5$---after thousands of iterations for the same setting as \fref{FIG:ALGCOMP:GT}. The visual impression of the images and kernels and the objective values indicate that all algorithms converge to a similar critical point.}\label{FIG:ALGCOMP:GT:ALGS}%
\end{figure}}%
{
\newcommand{\PlotAlg}[5]{
\PlotKernelSpy{#2, $\Obj$:#3}{\FilenameLocal__image_#1_parula.png}{\FilenameLocal__kernel_#1.png}{#4}{#5}{\SpyX}{\SpyY}}%
\begin{figure}[h]%
\def\SpyX{1.2cm}\def\SpyY{1.9cm}
\def\FilenameLocal{trees2_ch108_NW__PALM0__lambda_u__1__lambda_k_10__gamma_0-9995__eta_0-003}
\centering%
\ifexternalized%
    \includegraphics{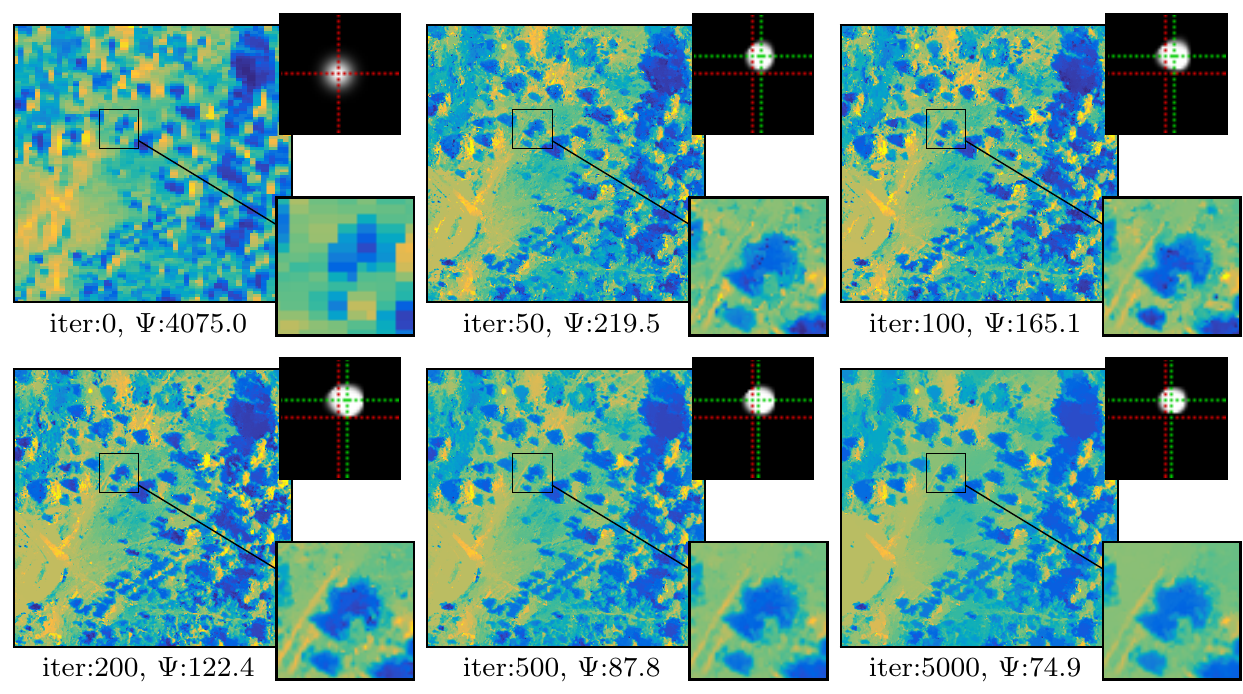}\\[-3mm]
\fi
\caption{Iterates of PALM for the data set \DataTreesA~with $\lambda_u = 1$ and $\lambda_k = 10$ with the objective function values given in brackets.} \label{FIG:ALGCOMP:TREES}%
\vspace*{3mm}
\ifexternalized%
    \includegraphics{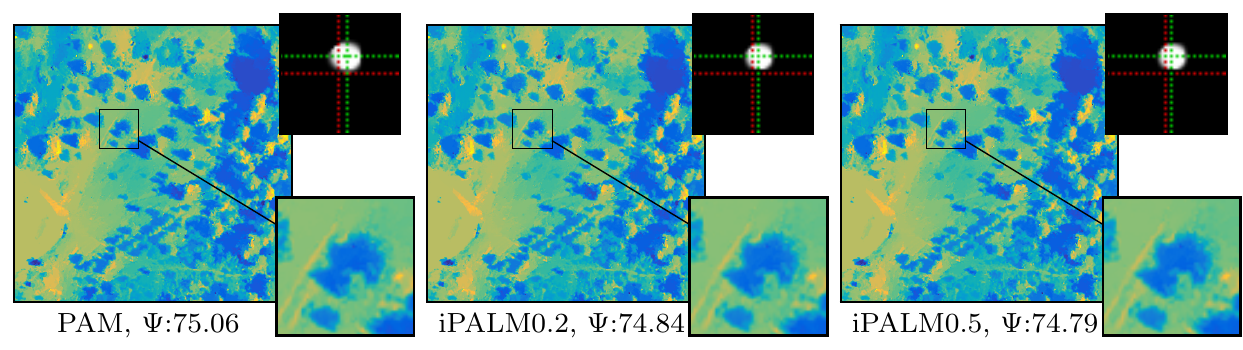}\\[-3mm]
\fi
\caption{Reconstructions by PAM and iPALM with inertial parameters $\alpha = 0.2$ and $\alpha=0.5$ for the setting of \fref{FIG:ALGCOMP:TREES}. Visually, the images and kernels and quantitatively, the objective function values indicate that all algorithms converge to the same critical point.}\label{FIG:ALGCOMP:TREES:ALGS}
\end{figure}}

{
\newcommand{\DrawStatsLocal}[1]{\includegraphics[height=3.2cm]{\FilenameLocal__#1.png}}%
\begin{figure}[h]%
\def\FilenameLocal{trees1_shift_5px_disk__lambda_u__0-1__lambda_k_10__gamma_0-9995__eta_0-003}
\centering%
\DrawStatsLocal{ObjectiveDecay}%
\DrawStatsLocal{ObjectiveTracking}%
\DrawStatsLocal{Legend}\\[-1mm]%
\DrawStatsLocal{DataFidelityTracking}%
\DrawStatsLocal{RuTracking}%
\DrawStatsLocal{RkTracking}\\[-3mm]%
\caption{Algorithm statistics over the course of the iterations for the same setting as \fref{FIG:ALGCOMP:GT}. The relative objective function values (\textbf{top left}) show that all algorithms have about the same speed on this example. The objective function values (\textbf{top center}) and its summands (\textbf{bottom}) indicate that also all algorithms converge to the same critical point. Overall, the plots show that more inertia leads to slower convergence.}\label{FIG:ALGCOMP:GT:STATS}
\vspace*{3mm}
\def\FilenameLocal{trees2_ch108_NW__lambda_u__1__lambda_k_10__gamma_0-9995__eta_0-003}
\centering%
\DrawStatsLocal{ObjectiveDecay}%
\DrawStatsLocal{ObjectiveTracking}%
\DrawStatsLocal{Legend}\\[-1mm]%
\DrawStatsLocal{DataFidelityTracking}%
\DrawStatsLocal{RuTracking}%
\DrawStatsLocal{RkTracking}\\[-3mm]%
\caption{Algorithm statistics for PALM and iPALM with a variety of inertial parameters for the same setting as \fref{FIG:ALGCOMP:TREES}. The objective function values (\textbf{top center}) and its summands (\textbf{bottom}) indicate that all algorithms converge to the same critical point. The relative objective function values (\textbf{top left}) show that for this example a higher inertial parameter leads to faster convergence.} \label{FIG:ALGCOMP:TREES:STATS}
\end{figure}}%
Before we focus on the actual image fusion reconstructions, we compare the algorithms PALM, iPALM and PAM on two different data sets. The results are shown in figures \ref{FIG:ALGCOMP:GT}, \ref{FIG:ALGCOMP:GT:ALGS}, \ref{FIG:ALGCOMP:GT:STATS} for the data set \DataGTdisk~and in figures \ref{FIG:ALGCOMP:TREES}, \ref{FIG:ALGCOMP:TREES:ALGS}, \ref{FIG:ALGCOMP:TREES:STATS} for \DataTreesA. Please refer to the captions of each figure for some detailed observations. We would like to highlight two important observations. First, both examples show that all three algorithms converge to practically the same critical point and that a visually pleasing reconstruction is already obtained after about 500 iterations. Thus, even though problem \eref{EQ:BID} is non-convex, the algorithm itself does not seem to have much influence on the final reconstructions. Second, the examples show that inertia may slow the convergence down as for \DataGTdisk~or may make it faster as for \DataTreesA. Other results---which are not shown here for brevity---indicate that this is correlated with the regularization parameter $\lambda_u$. For large values, e.g. $\lambda_u \geq 1$, we found that inertia consistently sped up convergence while for smaller values, e.g. $\lambda_u \leq 0.1$ we found that it was always slowing down the convergence. Thus, for simplicity and due to the lack of supporting examples that would justify other decisions, all further reconstructions are performed by PALM.

\subsection{Simulated Data}
{%
\begin{figure}%
\centering
\ifexternalized%
    \includegraphics{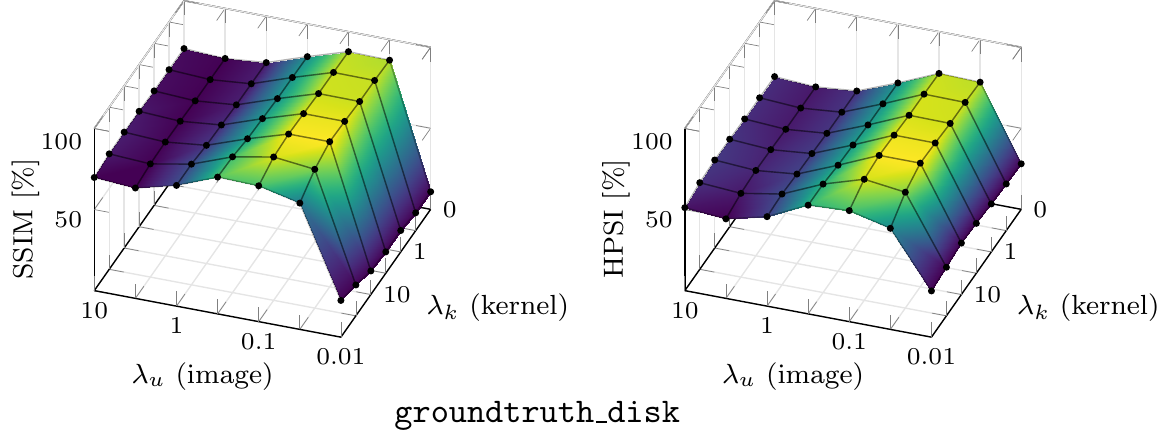}\\[3mm]%
    \includegraphics{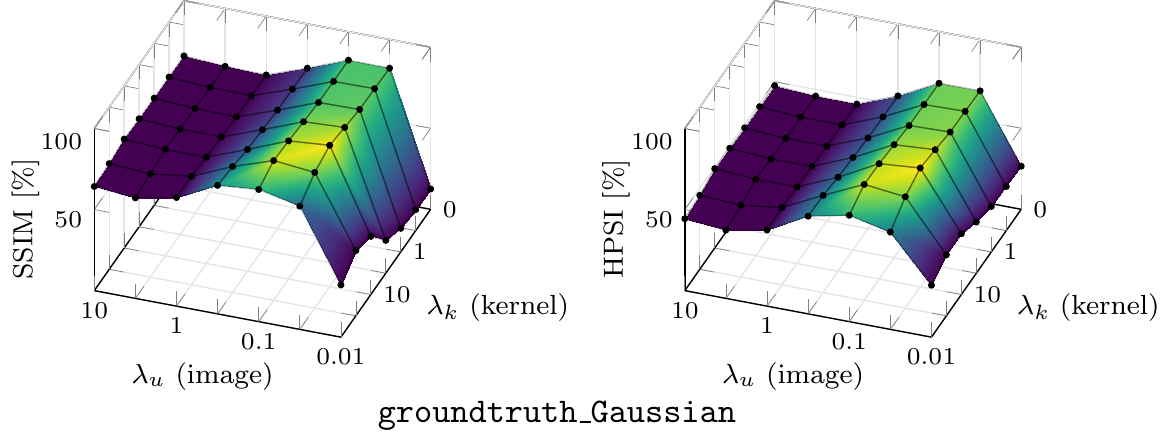}\\[-3mm]
\fi
\caption{Similarity measures SSIM and HPSI for different values of the regularization parameters $\lambda_u$ and $\lambda_k$. \textbf{Top:} data set \DataGTdisk ~with shifted side information. \textbf{Bottom:} data set \DataGTgaussian. Yellow corresponds to high and blue to low similarity. All four plots indicate that the reconstructions are not very sensitive with respect to the kernel regularization.} \label{FIG:GT_RegParam-u_RegParam-k}
\end{figure}}

{
\newcommand{\PlotSideInfoShifted}[2]{%
\draw (#1-0.04cm,#2-0.04cm) node [anchor=south west] {\fbox{\includegraphics[width=\PicWidth]{trees1_shift_5px_disk_side_info.png}}};%
\PlotSpy{side info}{trees1_shift_5px_side_info.png}{#1}{#2}{\SpyX}{\SpyY}}

\newcommand{\PlotLocal}[7]{
\PlotKernelSpy{#1}{#2}{#3}{#4}{#5}{\SpyX}{\SpyY}%
\draw (#4+.1cm, #5-.5cm) node [anchor=west] {\footnotesize SSIM: #6, HPSI: #7};}%

\begin{figure}%
\centering
\ifexternalized
    \includegraphics{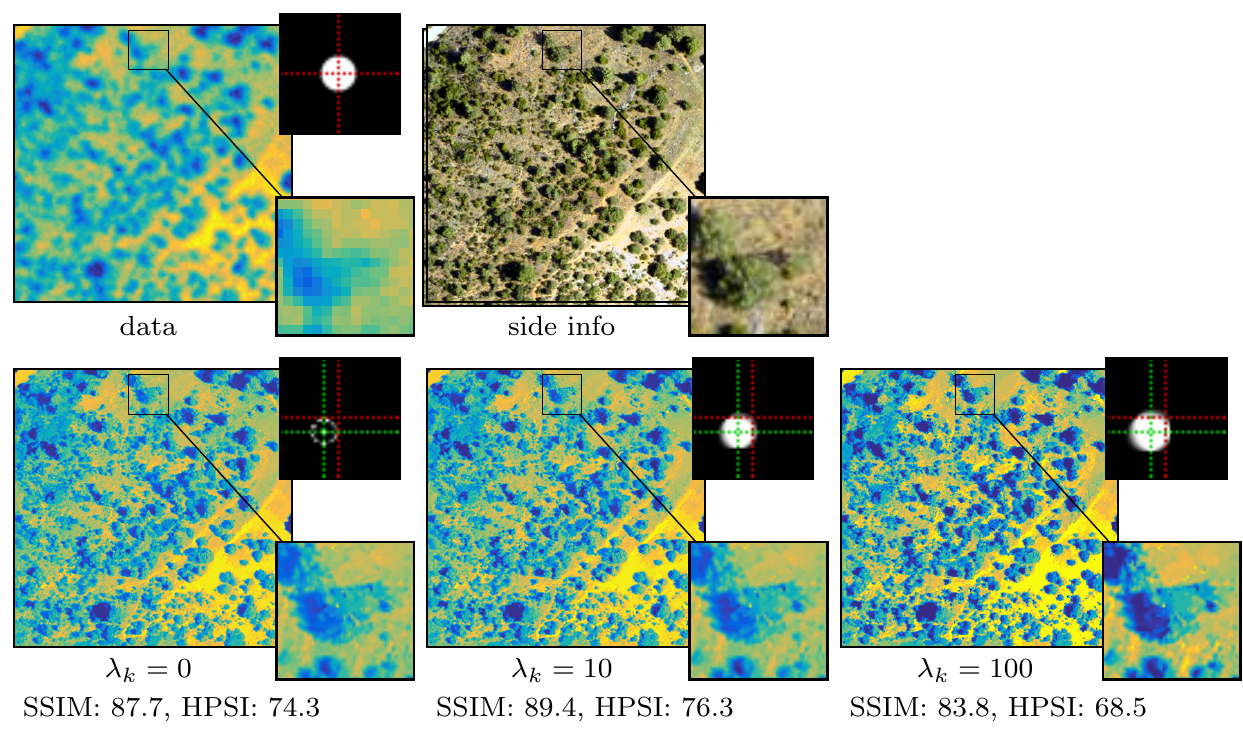}
\fi
\caption{Reconstructions for \DataGTdisk~with shifted side information. \textbf{Top:} Low resolution data and side information. \textbf{Bottom:} Reconstruction with varying image kernel regularization and image regularization chosen as $\RegParam_u=0.1$. SSIM and HPSI are given in percentages and higher values indicating better performance.}\label{FIG:ground_truth_results_disk}%
\vspace*{3mm}
\ifexternalized
    \includegraphics{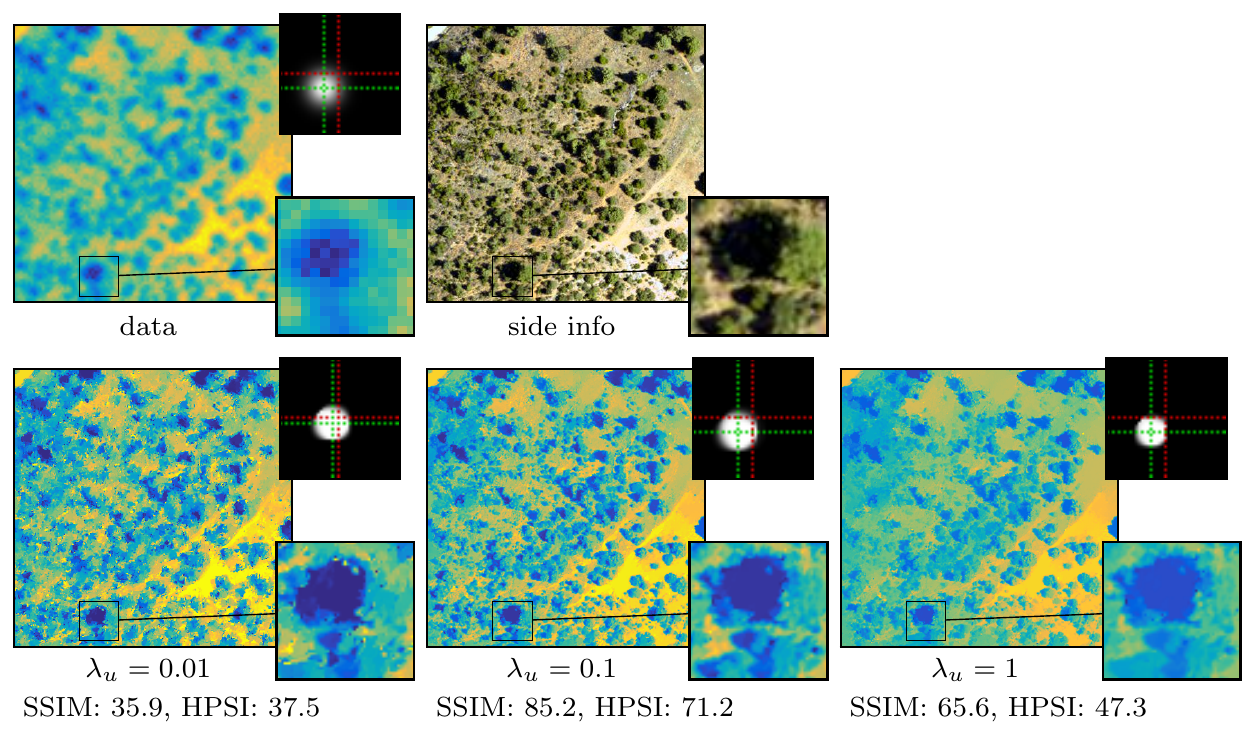}\\[-3mm]
\fi
\caption{Reconstructions for \DataGTgaussian. \textbf{Top:} Low resolution data and side information. \textbf{Bottom:} Reconstruction for varying image regularization and kernel regularization being chosen as $\RegParam_k = 10$. SSIM and HPSI are given in percentages and higher values indicating better performance.}\label{FIG:ground_truth_results_gauss}%
\end{figure}}

This section is dedicated to simulated data where the ground truth image is known. \Fref{FIG:GT_RegParam-u_RegParam-k} shows the similarity indices SSIM and HPSI of the reconstructed images when varying the image and kernel regularization parameters $\RegParam_u$ and $\RegParam_k$. The results show that the proposed model is stable with respect to the choice of parameters as long as the image regularization $\RegParam_u$ is not too small. It is remarkable that the success of the model depends only mildly on the choice of $\RegParam_k$ and a vanishing kernel regularization $\RegParam_k=0$ has only modest impact. However, it can also be seen that best reconstructions are obtained for non-zero $\RegParam_k$ which indicates that a mild kernel regularization should be performed.

In \fref{FIG:ground_truth_results_disk} we show reconstruction results for the \DataGTdisk~test case where we fix the image regularizing parameter $\RegParam_u=0.1$ and vary the kernel regularization $\RegParam_k$. Note that we complicated the situation by using a side information which is shifted by several pixels and, thus, \emph{not aligned} to the data. The proposed approach is well-fit to deal with this problem since blind deconvolution can implicitly perform rigid registration of data and side information by producing a blurring kernel that compensates for the translation by having an off-centered centroid. The results show that our method is able to compensate for unaligned side information images and can achieve very good results, nevertheless. The influence of the amount of kernel regularization is noticeably small although over-regularization clearly leads to different contrasts in the reconstructed image. Note that the barycentric translations of the kernels in \fref{FIG:ground_truth_results_disk} correspond exactly---up to pixel accuracy---to the number of pixels that the shifted side information deviates from the un-shifted one.%

\Fref{FIG:ground_truth_results_gauss} shows in a similar way the reconstruction results for the test case \DataGTgaussian, using the un-shifted side information. In contrast to the previous example, here we fix the parameter $\RegParam_k=10$ and vary $\RegParam_u$ instead. Here, the effects of different regularization strengths are considerably larger; particularly, in the case $\RegParam_u=0.01$ the shift between data and side information---which was induced by convolving with the off-centered kernel---is not fully detected. Consequently, the resulting similarity values, being not translationally invariant, are relatively poor. In the opposite scenario of over-regularizing the image, small structures in the image are not captured well.

\subsection{Hyperspectral Data}

{%
\newcommand{\PlotLocal}[5]{%
    \PlotKernelSpy{#1}{trees2_ch108_NW__PALM0__lambda_u__#2__lambda_k_#3__gamma_0-9995__eta_0-003_niter_2000_image_parula.png}{trees2_ch108_NW__PALM0__lambda_u__#2__lambda_k_#3__gamma_0-9995__eta_0-003_niter_2000_kernel.png}{#4}{#5}{\SpyX}{\SpyY}}
\newcommand{\DrawVLabel}[3]{\node[anchor=center] at (#2 - 0.1cm, #3 + 0.5 * \PicWidth) {\rotatebox{90}{\footnotesize\centering#1}};}
\begin{figure}%
\centering
\ifexternalized
    \includegraphics{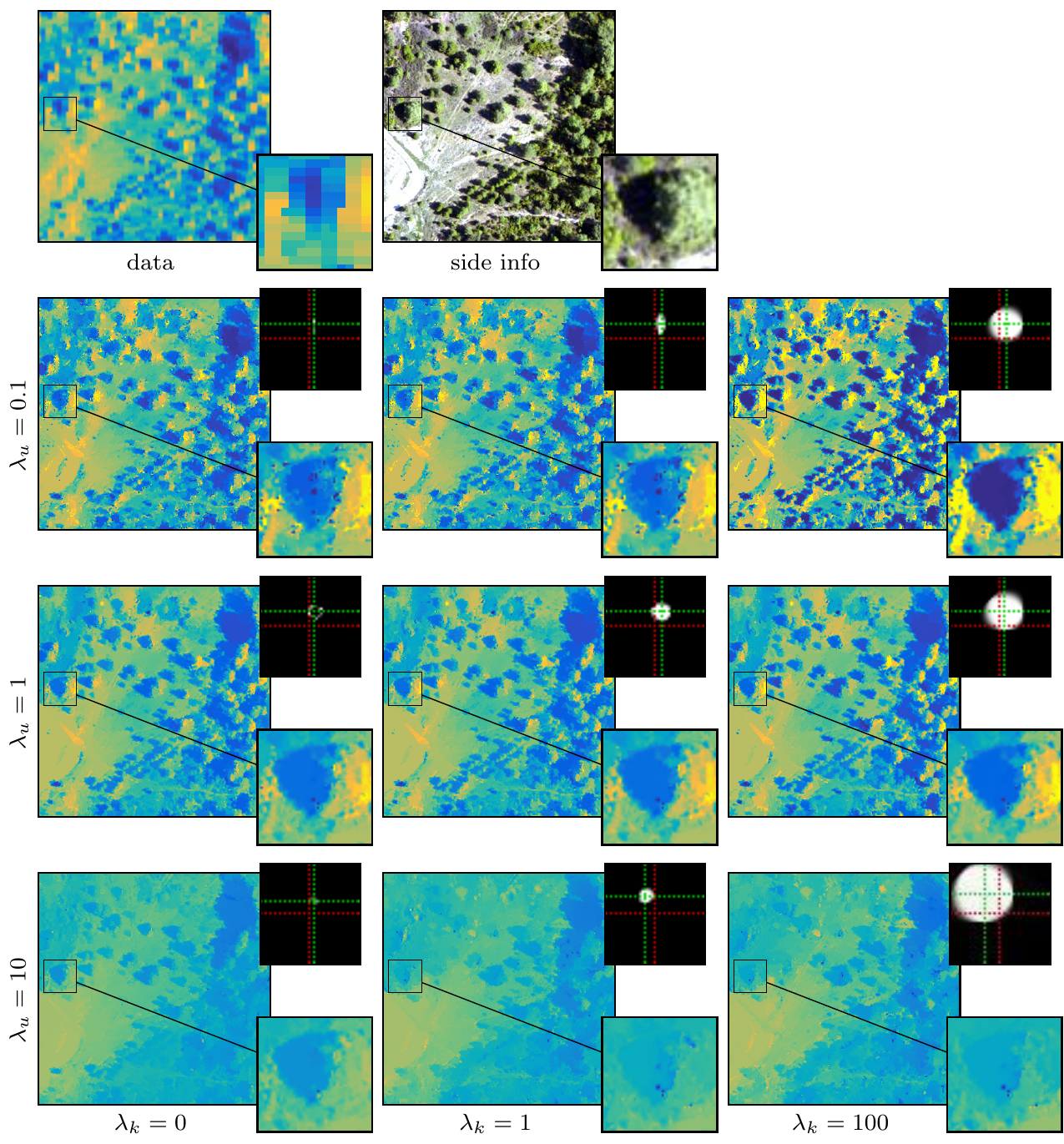}
\fi
\caption{Varying both regularization parameters for \DataTreesA~which corresponds to a near-infrared channel with wavelength of around 900 nm. Too small image regularization $\lambda_u$ leads to point artefacts and too large $\lambda_u$ to a loss of contrast and wrong estimation of the kernel. While no kernel regularization does not seem to have negative effects on the reconstruction, too large kernel regularization changes the image contrast.}\label{FIG:REGPARAM:CHANNEL108}%
\end{figure}}%

{%
\newcommand{\PlotLocal}[5]{%
    \PlotKernelSpy{#1}{urban_ch1_city__PALM0__lambda_u__#2__lambda_k_#3__gamma_0-9995__eta_0-003_niter_2000_image_parula.png}{urban_ch1_city__PALM0__lambda_u__#2__lambda_k_#3__gamma_0-9995__eta_0-003_niter_2000_kernel.png}{#4}{#5}{\SpyX}{\SpyY}}
\newcommand{\DrawVLabel}[3]{\node[anchor=center] at (#2 - 0.1cm, #3 + 0.5 * \PicWidth) {\rotatebox{90}{\footnotesize\centering#1}};}
\begin{figure}%
\centering
\ifexternalized
    \includegraphics{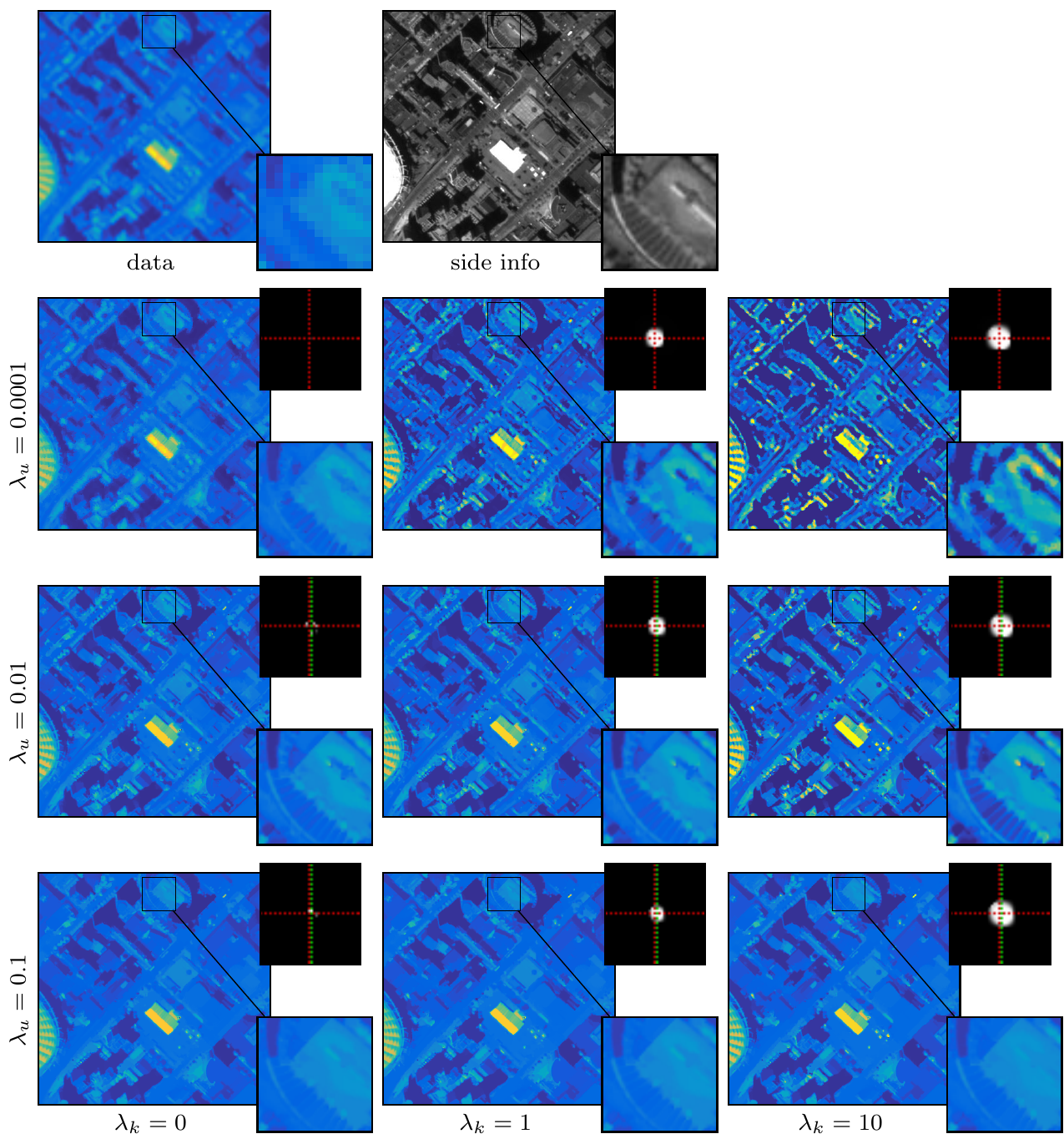}
\fi
\caption{Results for \DataUrbanA~(near-infrared) for varying image and kernel regularization. Over- and underregularization have similar effects like in \fref{FIG:REGPARAM:CHANNEL108}. However, due to a lower noise level of the data, the parameters can be chosen smaller.}
\label{FIG:REGPARAM:CHANNEL1}%
\end{figure}}%

\begin{figure}[h]%
\centering%
\ifexternalized
    \includegraphics{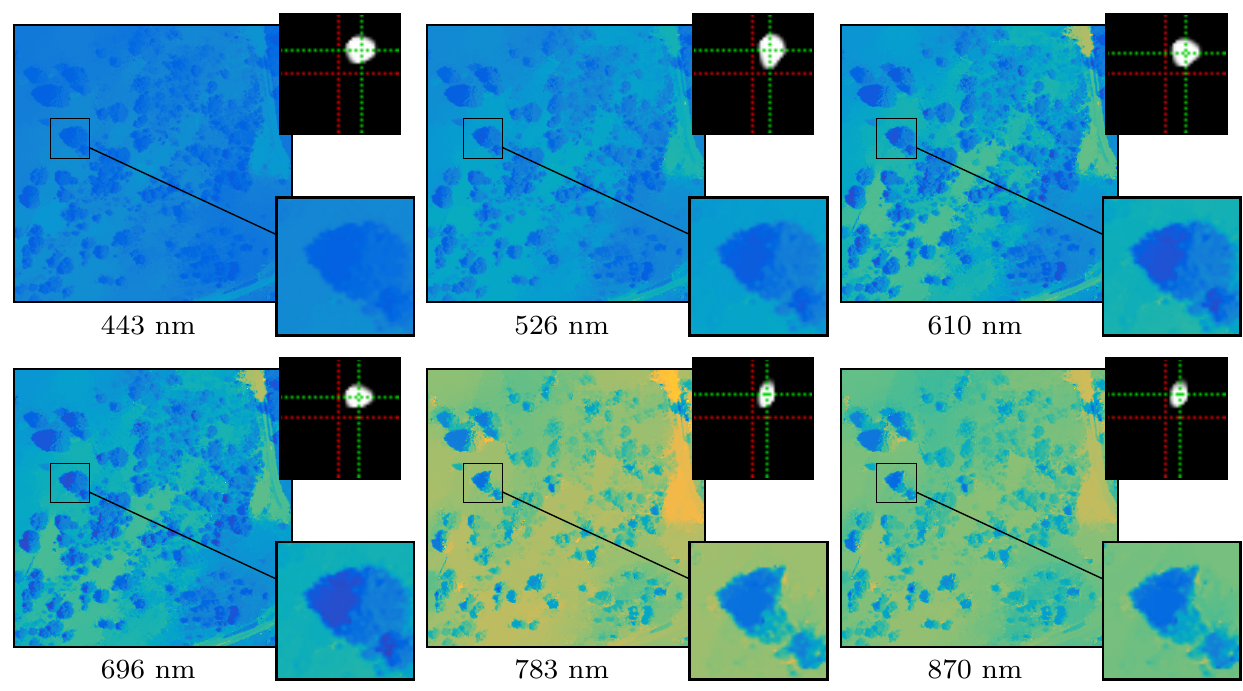}
\fi
\caption{Reconstructions of several channels corresponding to different wavelengths given in nanometer (nm) of \DataTreesB. The regularization parameters for this experiment were chosen constant across the channels as $\RegParam_u = 1$ and $\RegParam_k = 1$. For visualization, reconstructed images are scaled back to their physical range. Note that despite the reconstruction being performed channel-by-channel, all kernels have about the same shape and the same shift.}\label{FIG:DATA1:CHANNELS}
\vspace*{3mm}
\centering%
\ifexternalized
    \includegraphics{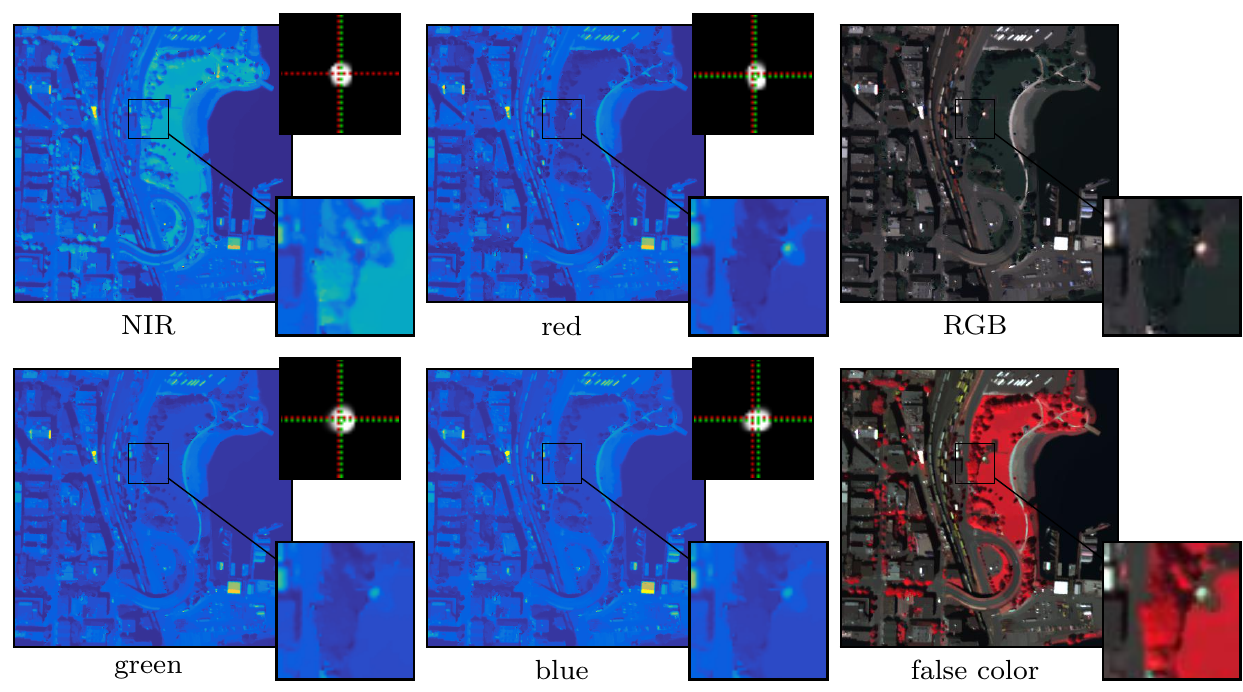}
\fi
\caption{Reconstruction of the multispectral data set \DataUrbanB. The regularization parameters were chosen the same for all channels as $\RegParam_u=0.01$ and $\RegParam_k=1$. Kernels are being estimated consistently over the range of the channels up to an accuracy of one pixel. The resulting RGB and false color images have well-defined boundaries and show no color smearing. (NIR = near-infrared)} \label{FIG:CITY:CHANNELS}
\end{figure}

In this section we apply our method to real data and---similarly to the previous section--- investigate the effects of varying image and kernel regularization. To this end, we apply our method to one channel of the hyperspectral environmental and of the multispectral urban data set.  Figures \ref{FIG:REGPARAM:CHANNEL108} and \ref{FIG:REGPARAM:CHANNEL1} show data, side information, and the corresponding reconstruction results for different values of the regularization parameters. Since no ground truth solutions are known in this case, we rely on our visual impression to choose a set of parameters. \Fref{FIG:REGPARAM:CHANNEL108}, which corresponds to near-infrared light, suggests that a good compromise among data fidelity and regularization is achieved for $\RegParam_u=\RegParam_k=1$. However, even if no $\tv$ regularity of the kernel is required, that is in the case $\RegParam_k=0$, the resulting images do hardly differ. Bearing in mind the results of \fref{FIG:ground_truth_results_disk}, we will regularize the kernel slightly in the sequel. The situation is similar in \fref{FIG:REGPARAM:CHANNEL1}, which also corresponds to near-infrared light. However, a smaller image regularization parameter, e.g. $\RegParam_u=0.01$, seems necessary in order to avoid over-smoothing. %

Finally, we apply the proposed strategy to several hyperspectral channels of the data set \DataTreesB, using the same parameters for the reconstruction of each channel. Here, we scaled the reconstructed images back to their physical range to allow inter-channel comparisons. \Fref{FIG:DATA1:CHANNELS} shows the reconstruction of six hyperspectral channels ranging from blue to near-infrared light. Note that the resulting kernels only change slightly with respect to the channel. We would like to point out that we were able to use the same set of parameters, namely $\RegParam_u=\RegParam_k=1$, for all channels. In \fref{FIG:CITY:CHANNELS} we show the reconstruction of all channels of the multispectral urban data set \DataUrbanB~using $\RegParam_u=0.01$ and $\RegParam_k=1$. Additionally, we visualized the results by re-combining the four channels into an RGB and a false color image which has the channels near-infrared, red and green. We would like to highlight that the kernels are consistently estimated up to a pixel accuracy and that no color smearing is visible in the RGB and false color images despite the fact that all channels have been processed independently.%

\section{Conclusions and Outlook} We have presented a novel model for simultaneous image fusion and blind deblurring of hyperspectral images based on the directional total variation. We showed that this approach yields sharp images for both environmental and urban data sets that are either acquired by a plane or by a satellite. In addition, the numerical comparison of different algorithms showed that---despite the non-convexity of the problem---the computed solutions are stable with respect to the algorithms. The implicitly included registration by estimating the kernel yields further robustness to imperfections in the image acquisitions. Thus, the proposed approach is appealing for automated processing of large data sets. 

Future work will be devoted to multiple directions. First, the proposed approach is channel-by-channel which makes it computationally appealing. However, a coupling of the spectral channels via their kernel and/or via spectral fingerprint preservation may yield further improvements. Second, higher order regularization such as the total generalized variation \cite{Bredies2010} has been proven useful for natural images. This naturally leads to the directional total generalized variation which combines higher order smoothness with structural a-priori knowledge as presented in this work, thereby generalizing the definition of \cite{Kongskov2017}. Similarly, other kernel regularization functionals such as the total generalized variation or the $H^1$-semi-norm will be considered in future investigations to better model the smoothness of the kernel. \rev{Finally, it is currently not clear how the computational cost of the proposed method depends on the size of the considered data. In particular, it is difficult to predict to which degree larger images also require a higher number of iterations during optimization for the method to yield satisfactory results. However, due to the fact that our approach mostly depends on operations that are of a local nature, it should be highly suitable for parallelization. The development and analysis of implementations that utilize the benefits of distributed computing architectures is also a topic for future research.}

\ack
We would like to express our gratitude to Juheon Lee and Qi Wei for their help with the environmental and urban data sets, respectively. Furthermore, the authors would like to thank Deimos Imaging for acquiring and providing the data used in this study, and the IEEE GRSS Image Analysis and Data Fusion Technical Committee.

M.J.E. and C.-B.S. acknowledge support from Leverhulme Trust project \enquote{Breaking the non-convexity barrier}, EPSRC grant \enquote{EP/M00483X/1}, EPSRC centre \enquote{EP/N014588/1}, the Cantab Capital Institute for the Mathematics of Information, and from CHiPS (Horizon 2020 RISE project grant). Moreover, C.-B.S. is thankful for  support from the Alan Turing Institute.    

D.A.C. acknowledges the support of NERC (grant number NE/K016377/1). We are grateful to NERC’s Airborne Research Facility and Data Analysis Node \url{https://arsf-dan.nerc.ac.uk/} for conducting the airborne survey and pre-processing the environmental data collected from  Alto Tajo, Spain (survey CAM11/03).

\appendix%
\setcounter{section}{1}%
\section*{Appendix}
\begin{figure}[h!]
\centering
\ifexternalized
    \includegraphics{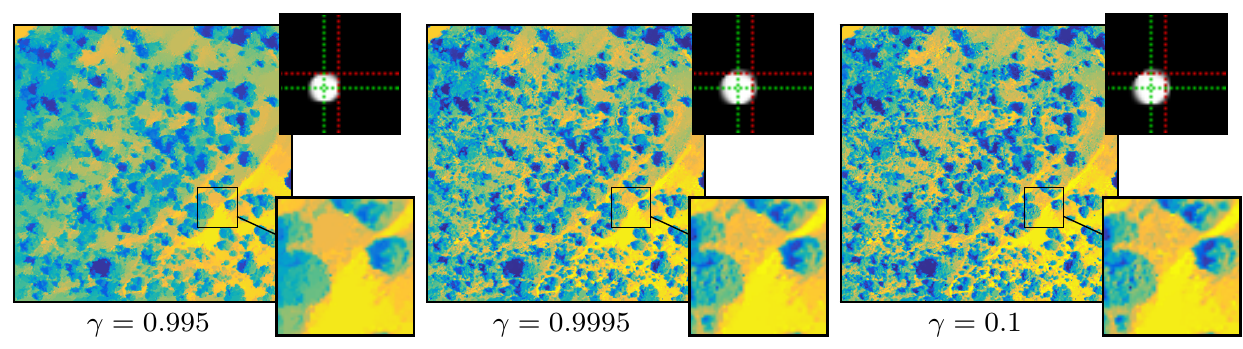}\\%
    \includegraphics{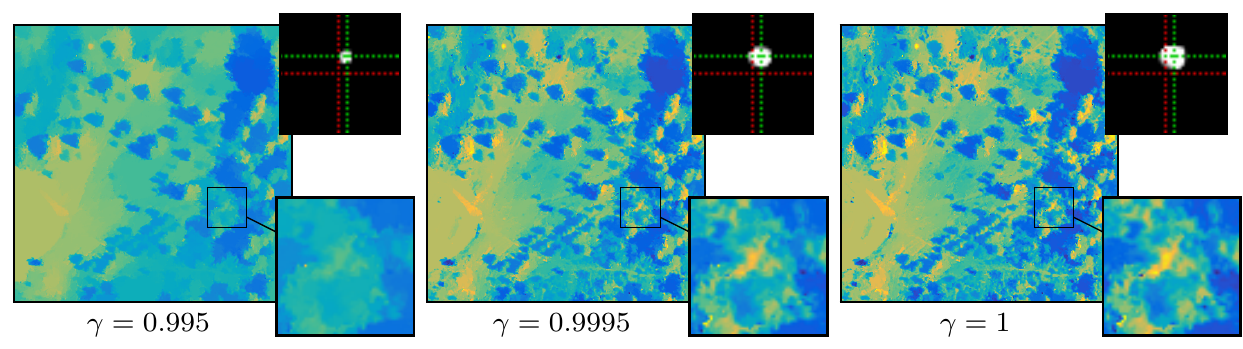}
\fi
\caption{Reconstructions for varying vector field parameters $\gamma$. \textbf{Top:} Results for \DataGTdisk~with $\RegParam_u=0.1$ and $\RegParam_k=10$. \textbf{Bottom:} Results for \DataTreesA~with $\RegParam_u=1$, $\RegParam_k=1$.}\label{FIG:GAMMA}%
\end{figure}

\rev{We test the influence of parameter $\gamma$ in \eref{EQ:vectorfield} on a simulated and a real data set. \Fref{FIG:GAMMA} shows the corresponding reconstruction results for $\gamma\in\{0.995,0.9995,1\}$ where the middle value is used in all other numerical experiments in this work. Note that the results for $\gamma=1$ suffer from stronger point artifacts whereas those for $\gamma=0.995$ do not capture local structures well enough.}

\bibliography{bibliography}
\end{document}